\documentclass[sigconf]{acmart}
\AtBeginDocument{%
  }

\setcopyright{acmlicensed}
\copyrightyear{2018}
\acmYear{2018}
\acmDOI{XXXXXXX.XXXXXXX}
\acmConference[Conference acronym 'XX]{Make sure to enter the correct
  conference title from your rights confirmation email}{June 03--05,
  2018}{Woodstock, NY}
\acmISBN{978-1-4503-XXXX-X/2018/06}

\usepackage{multirow} 
\usepackage{caption}
\usepackage{graphicx}
\usepackage{subcaption} 
\usepackage{amsmath}
\usepackage{algorithm}
\usepackage{algorithmic}
\usepackage{tikz}
\usepackage{comment}
\usepackage{derivative}
\usepackage{makecell}
\usepackage{booktabs}
\usepackage{pifont}   
\usepackage{xcolor}
\usepackage{caption}

\newcommand{\meanstd}[2]{#1 {\tiny (#2)}}

\copyrightyear{2026}
\acmYear{2026}
\setcopyright{cc}
\setcctype{by}
\acmConference[CIKM '26]{Proceedings of the 35th ACM International Conference on Information and Knowledge Management}{November 07--11, 2026}{Rome, Italy}
\acmBooktitle{Proceedings of the 35th ACM International Conference on Information and Knowledge Management (CIKM '26), November 07--11, 2026, Rome, Italy}
\acmDOI{10.1145/3799682.3840917}
\acmISBN{979-8-4007-2539-5/2026/11}

\begin{document}

\title{VisAdj: Learning Adjacency Matrices from Node-Link Images}

\author{Jiahao Xie}
\email{jiahaox@udel.edu}
\orcid{0009-0002-8716-6566}
\affiliation{%
  \institution{University of Delaware}
  \city{Newark}
  \state{Delaware}
  \country{USA}
}

\author{Guangmo Tong}
\email{amotong@udel.edu}
\orcid{0000-0003-3247-4019}
\affiliation{%
  \institution{University of Delaware}
  \city{Newark}
  \state{Delaware}
  \country{USA}
}

\renewcommand{\shortauthors}{Jiahao Xie and Guangmo Tong}

\begin{abstract}
Learning adjacency matrices from node-link images is a fundamental problem for recovering structured graph information from visual observations. 
Existing methods typically rely on fixed KNN-based heuristics for candidate edge selection and fail to capture dependencies among edges.
To overcome these limitations, we propose VisAdj, a new framework for topology-aware adjacency prediction. VisAdj introduces an attention-sparse neighbor sampler to adaptively select a high-recall set of candidate node pairs and performs joint edge inference using a line-graph transformer that treats candidate edges as tokens and explicitly models dependencies among incident edges.
Extensive experiments on synthetic graphs, road networks, and vessel images demonstrate that VisAdj consistently outperforms existing baselines by clear margins. 

\end{abstract}



\maketitle

\section{Introduction}
\begin{figure}[t!]
    \centering 
    \subfloat[Road Network]{\label{fig:road_network}\includegraphics[width=0.15\textwidth]{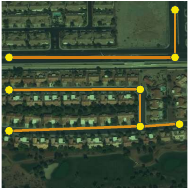}}\hspace{1.5mm}
    \subfloat[Vessel Network]{\label{fig:vessel_network}\includegraphics[width=0.15\textwidth]{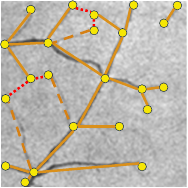}}\hspace{1.5mm}
    \subfloat[Graph Image]{\label{fig:graph_image}\includegraphics[width=0.15\textwidth]{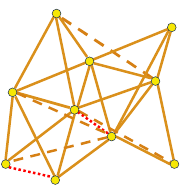}}
    \captionsetup{skip=6pt}
\caption{Examples of node-link images and their underlying graphs. 
Orange lines denote ground-truth edges. 
Existing methods often miss long-range connections (orange dashed lines) and predict erroneous shortcut edges (red dotted lines) in visually ambiguous crossing regions.
}
    \Description{Three example node-link images are shown: a road network, a vessel network, and a general graph image. Each image contains visible nodes and links that correspond to an underlying graph structure.}
    \label{fig:visualization_process}
\end{figure}

Node-link images are widely used to visualize relational structures \cite{saket2014node,jianu2014display,he2022td, xie2026vsal}, 
such as road and vessel networks (Fig.~\ref{fig:visualization_process}), 
where the underlying graphs encode the semantic information of interest and the images serve primarily for human interpretation.
However, in practice, the original graph data are often unavailable, leaving only rasterized images 
such as satellite photos and medical images. Consequently, the adjacency structures become inaccessible to downstream graph algorithms and learning models. This introduces the problem of recovering adjacency matrices from node-link~images. 

Early attempts recover graphs from node-link images using hand-crafted heuristics~\cite{das1997adjacency}, which are sensitive to visual variations. 
Recent road network extraction methods, such as SAM-Road \cite{hetang2024segment} and SAM-Road++~\cite{yin2025towards}, adopt learning-based pipelines but still rely on fixed KNN-based strategies to select candidate node pairs.
This leads to an inherent limitation: a small neighborhood radius misses long-range connections, whereas a large radius admits many spurious candidates. Such designs are effective for road networks with relatively simple structures and limited crossings (Fig.~\ref{fig:road_network}), but become less reliable for general node-link images, where the underlying graph structures may be more complex (Figs.~\ref{fig:vessel_network}-\ref{fig:graph_image}).
More general image-to-graph methods, such as RelationFormer~\cite{shit2022relationformer} and Any2Graph~\cite{krzakala2024any2graph}, infer graph structures through learned relational reasoning, but still predict each edge independently based only on endpoint features, without considering structural dependencies among edges.
In practice, edge dependencies come from both node-level and graph-level structural constraints: node degree limits the number of incident edges, and global graph properties (e.g., planarity in Fig.~\ref{fig:road_network} and tree topology in Fig.~\ref{fig:vessel_network}) impose additional topological restrictions. Ignoring such edge-edge interactions often results in ambiguous or inconsistent adjacency predictions, making edge inference the primary bottleneck in graph reconstruction.

\textbf{Contribution.}
We propose \textbf{VisAdj}, a new framework for reconstructing adjacency matrices from node-link images. VisAdj first encodes the input image into local and global visual features to support node detection and image-level structural understanding, with global features enhanced by learnable topology tokens. 
To overcome the rigidity of KNN-based candidate selection, VisAdj introduces an attention-sparse neighbor sampler to adaptively select a high-recall set of candidate node pairs. 
Furthermore, to address the limitation that existing methods ignore edge dependencies, VisAdj performs joint edge inference over candidate node pairs using a line-graph transformer, which treats edges as tokens and explicitly models interactions among incident edges for structurally consistent adjacency prediction.
Experimentally, VisAdj improves the state of the art by over $15\%$ in graph isomorphism rate and over $8\%$ in edge prediction F1 score across diverse benchmarks, while also enhancing existing road network extraction pipelines as a plug-in graph reasoning module.
Our code, dataset, and hyperparameter settings are available at 
\url{https://github.com/Jiahao-Xie-86/VisAdj}.
\clearpage

\section{Related Work}
\textbf{Road network extraction}. A large body of work studies image-to-graph reconstruction in the context of road network extraction~\cite{mattyus2017deeproadmapper,bastani2018roadtracer, tan2020vecroad, lu2025deep}. 
Sat2Graph~\cite{he2020sat2graph} models road graphs by predicting connectivity between detected junctions using convolutional features over local neighborhoods. RNGDet++~\cite{xu2023rngdet++} adopts an iterative graph-growing paradigm, progressively adding nodes and edges through region-of-interest sampling guided by historical maps. More recent approaches, such as SAM-Road \cite{hetang2024segment} and SAM-Road++ \cite{yin2025towards}, leverage vision foundation models to produce node and edge representations, which are subsequently converted into graph structures via post-processing. 
Despite their effectiveness on road datasets, these methods struggle to generalize to general node-link images because their graph reasoning modules are tailored to road-specific characteristics (e.g., planarity and local connectivity) and rely on fixed KNN-based methods for candidate edge selection.

\textbf{General image-to-graph methods}. In addition to road networks, several methods aim to recover more general graph structures from images~\cite{berger2025cross}. For example, RelationFormer~\cite{shit2022relationformer} introduces a transformer-based framework that jointly predicts nodes and edges using set prediction and bipartite matching, enabling end-to-end graph inference without explicit post-processing. 
Any2Graph \cite{krzakala2024any2graph} improves this paradigm by proposing an optimal transport-based loss to better align predicted and ground-truth graphs under permutation ambiguity. 
These approaches move toward unified image-to-graph modeling by jointly learning node and edge prediction. However, they predict each edge independently and lack explicit mechanisms to model structural dependencies among edges. 
In contrast, our framework overcomes these limitations by learning edge interactions and performing joint edge reasoning.

\section{Preliminary}
We study the problem of recovering graph adjacency matrices from node-link images.
Let $\mathcal{I}$ denote the space of node-link images and $\mathcal{G}$ the space of undirected, unweighted graphs. 
Each image $\mathbf{I} \in \mathcal{I}$ provides a visualization of the underlying graph represented as
$G = (V, \mathbf{A}) \in \mathcal{G}$, where $V$ denotes the node set and
$\mathbf{A} \in \{0,1\}^{|V|\times|V|}$ is the corresponding adjacency matrix.
Given a dataset 
$D = \{(\mathbf{I}_i, G_i=(V_i, \mathbf{A}_i))\}_{i=1}^n$,
the goal is to learn a mapping $h : \mathcal{I} \rightarrow (\widehat{V}, \widehat{\mathbf{A}})$ that recovers the underlying adjacency structure. 
Since nodes are grounded in image space, node identities are defined by their spatial locations.
During training, predicted nodes are aligned with ground-truth nodes via spatial matching,
which induces a consistent ordering between $\widehat{V}$ and $V$. Under this correspondence, $h$ is learned by minimizing the empirical loss
\begin{align*}
\mathcal{L}(h)
=
\frac{1}{n}\sum_{i=1}^{n}
\Big[
\ell_{\text{node}}\!\big(\widehat{V}_h(\mathbf{I}_i),V_i\big)
+
\ell_{\text{edge}}\!\big(\widehat{\mathbf{A}}_h(\mathbf{I}_i),\mathbf{A}_i\big)
\Big],
\end{align*}
where $\ell_{\text{node}}$ measures the difference between predicted and ground-truth node sets, and $\ell_{\text{edge}}$ measures the discrepancy between predicted and ground-truth adjacency matrices under the induced node correspondence (e.g., cross-entropy loss).

\begin{figure*}
    \centering
    \includegraphics[width=1\linewidth]{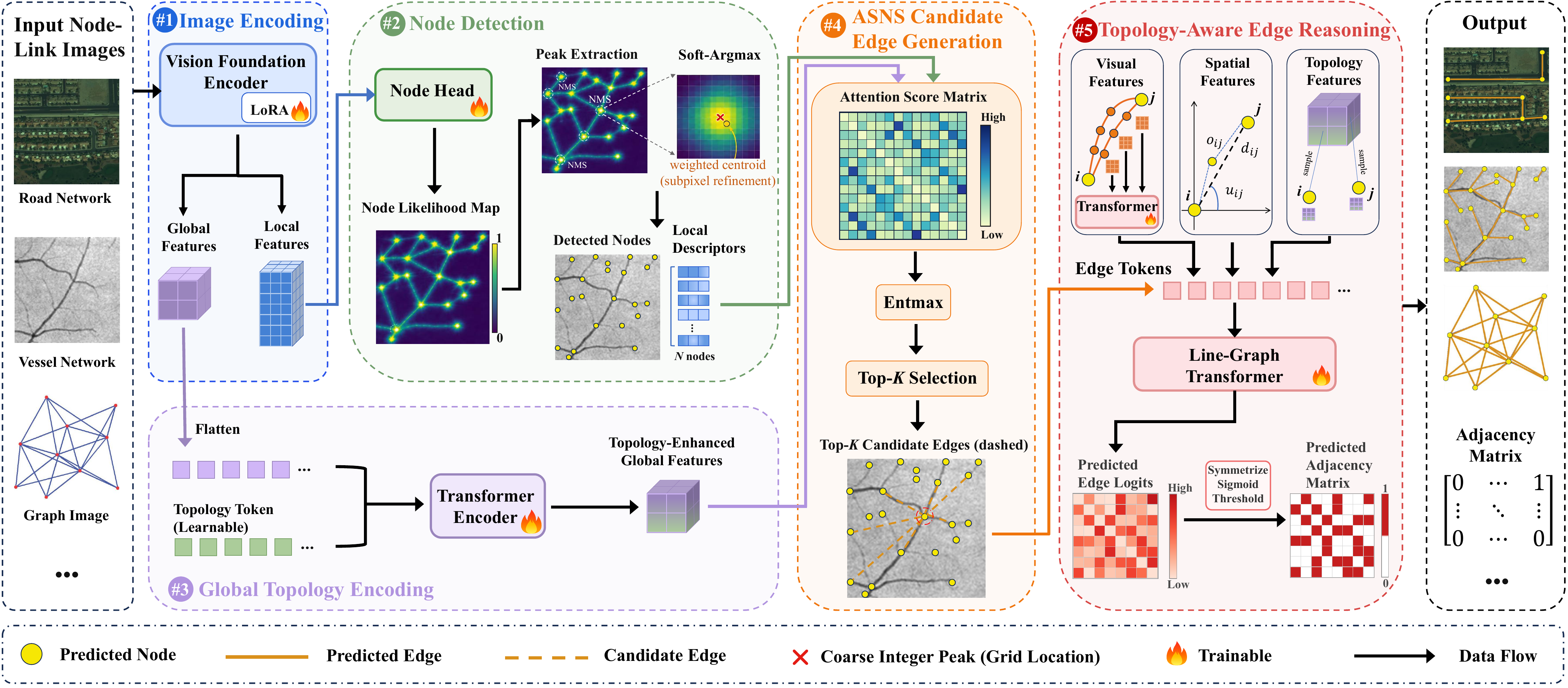}
    \caption{Overview of VisAdj. Given an input node-link image, VisAdj first (\#1) extracts local and global visual features using a vision foundation encoder, and (\#2) detects graph nodes from a node likelihood map through peak extraction, non-maximum suppression, and soft-argmax refinement. It then (\#3) enhances global features with learnable topology tokens to capture image-level structural context. Conditioned on the detected nodes, VisAdj (\#4) applies ASNS to generate a sparse high-recall set of candidate edges, and (\#5) represents these candidates as edge tokens for topology-aware reasoning with a line-graph transformer. The resulting edge logits are symmetrized and thresholded to obtain the predicted adjacency matrix.}
    \Description{Overview of the VisAdj architecture. The pipeline contains five numbered modules: image encoding, node detection, global topology encoding, ASNS candidate edge generation, and topology-aware edge reasoning with a line-graph transformer.}
    \label{fig:framework}
\end{figure*}

\section{Methodology} 
Given a node-link image $\mathbf{I}$, VisAdj aims to recover the underlying graph by predicting a node set  
and an adjacency matrix, i.e., $\big(\widehat{V}(\mathbf{I}),\, \widehat{\mathbf{A}}(\mathbf{I})\big)$. To this end, VisAdj follows a three-stage pipeline:
\textbf{1)} \textit{ Image encoding:} 
we encode $\mathbf{I}$ using a vision foundation model to obtain multi-scale features. 
\textbf{2)} \textit{Node detection:} the node set $\widehat{V}(\mathbf{I})$ is detected via node likelihood prediction.
\textbf{3)} \textit{Edge reasoning:} 
based on $\widehat{V}(\mathbf{I})$, VisAdj aggregates image-level context with learnable topology tokens, adaptively selects candidate node pairs using an attention-sparse neighbor sampler (ASNS), and jointly infers connectivity by employing a line-graph transformer to model dependencies among edges and predict $\widehat{\mathbf{A}}(\mathbf{I})$.
The overall architecture of our framework is illustrated in Fig.~\ref{fig:framework}, where the numbered blocks correspond to the main modules of VisAdj.
In the following, we first detail the architecture and then describe the training strategy.

\subsection{Architecture}

\subsubsection{Image Encoding}
Given an input node-link image $\mathbf{I} \in \mathbb{R}^{H\times W\times 3}$ with resolution $H \times W$, we extract visual representations using a pretrained vision foundation model \cite{awais2025foundation, liu2026dual} (e.g., SAM \cite{kirillov2023segment}, SAM2 \cite{ravi2025sam} and SAM3 \cite{carion2025sam}), 
as shown in block (\#1) of Fig.~\ref{fig:framework}.
The encoder $\mathcal{E}_{\boldsymbol{\theta}_{E}}$ first maps the input image to a backbone
feature representation, which is further projected into two complementary feature streams~\cite{carion2020end}:
\begin{align}
\nonumber
\mathbf{F}_{\text{local}}
&=
\mathcal{F}_{L,\boldsymbol{\theta}_{L}}
\big(\mathcal{E}_{\boldsymbol{\theta}_{E}}(\mathbf{I})\big)
\in \mathbb{R}^{H_L \times W_L \times D_L },
\\
\label{eq:global}
\mathbf{F}_{\text{global}}
&=
\mathcal{F}_{G,\boldsymbol{\theta}_{G}}
\big(\mathcal{E}_{\boldsymbol{\theta}_{E}}(\mathbf{I})\big)
\in \mathbb{R}^{H_G \times W_G \times D_G},
\end{align}
where $\boldsymbol{\theta}_{E}$ denotes the parameters of the vision backbone. 
$(H_L, W_L)$ and $(H_G, W_G) = (H_L/\lambda_G, W_L/\lambda_G)$ denote the spatial resolutions of the local and global feature streams, respectively, and $\lambda_G \in \mathbb{Z}^+$ is the downsampling ratio. $D_L, D_G \in \mathbb{Z}^+$ are their corresponding feature dimensions.
Both $\mathcal{F}_{L,\boldsymbol{\theta}_{L}}$ and $\mathcal{F}_{G,\boldsymbol{\theta}_{G}}$ are 
learnable feature representation modules 
parameterized by $\boldsymbol{\theta}_{L}$ and $\boldsymbol{\theta}_{G}$ (e.g., CNN~\cite{he2016deep}).
The local feature stream retains higher spatial resolution and preserves fine-grained geometric cues that are important for accurate node localization. In contrast, the global stream aggregates features at a coarser spatial scale, enlarging the receptive field and capturing long-range context that is critical for subsequent edge reasoning.

\subsubsection{Node Detection}

We detect graph nodes by estimating a dense node-likelihood map from $\mathbf{F}_{\text{local}}$, followed by peak extraction for node localization and feature sampling to obtain node descriptors.
This node detection process is illustrated in block (\#2) of Fig.~\ref{fig:framework}.

\textbf{1) Node likelihood prediction}. We first predict a dense node-likelihood map over the image by applying a node prediction head $\mathcal{H}_{\text{node},\boldsymbol{\theta}_{N}}$ with parameters $\boldsymbol{\theta}_{N}$  to process $\mathbf{F}_{\text{local}}$:
\begin{align}
\label{eq:node_mask}
\widehat{\mathbf{M}}
=
\mathcal{U}_{H,W}
\left[
\sigma\left(
\mathcal{H}_{\mathrm{node},\boldsymbol{\theta}_N}
(\mathbf{F}_{\mathrm{local}})
\right)
\right]
\in [0,1]^{H\times W},
\end{align}
where $\sigma$ denotes the sigmoid function 
and $\mathcal{U}_{H,W}$ is bilinear upsampling to the original image resolution. 
The resulting node-likelihood map $\widehat{\mathbf{M}}$ assigns each pixel a probability of being a graph node, providing dense spatial evidence for node existence. In practice, $\mathcal{H}_{\text{node},\boldsymbol{\theta}_{N}}$ is implemented as a convolutional predictor composed of CoordConv layers~\cite{liu2018intriguing} followed by standard convolutions.

\textbf{2) Peak extraction with subpixel refinement.} 
Candidate nodes are extracted as local maxima in the predicted likelihood map. A pixel $(\tilde{x},\tilde{y})\in\mathbb{Z}^2$ is selected as a node candidate if
\begin{align*}
 \widehat{\mathbf{M}}[\tilde{x},\tilde{y}] = \max_{(u,v) \in \mathcal{N}_r(\tilde{x}, \tilde{y})} \widehat{\mathbf{M}}[u,v] \quad \text{and} \quad \widehat{\mathbf{M}}[\tilde{x}, \tilde{y}] > \tau,
\end{align*}
where $\mathcal{N}_r(\tilde{x}, \tilde{y})$ denotes an $r\in\mathbb{R}^+$-radius neighborhood around the pixel $(\tilde{x},\tilde{y})$, and $\tau \in [0,1]$ is a confidence threshold. We apply non-maximum suppression (NMS) \cite{hosang2017learning} to merge spatially nearby detections that correspond to the same node.
To improve localization accuracy, we refine each detected peak using a differentiable soft-argmax operation. For a peak located at integer coordinates $(\tilde{x}, \tilde{y})$, the subpixel-refined node coordinates are computed via
\begin{align*}
(x,y) =
\sum_{(u',v')\in\mathcal{N}_w(\tilde{x},\tilde{y})}
\frac{\exp\big(\widehat{\mathbf{M}}[u',v']/T\big)\cdot (u', v')}
{\sum_{(u,v)\in\mathcal{N}_w(\tilde{x},\tilde{y})} \exp\big(\widehat{\mathbf{M}}[u,v]/T\big)},
\end{align*}
where $\mathcal{N}_w(\tilde{x},\tilde{y})$ denotes a local neighborhood of radius $w\in\mathbb{R}^+$ and $T\in\mathbb{R}^+$ is a temperature parameter controlling the sharpness of the localization. This procedure aggregates evidence from multiple pixels and yields node coordinates with continuous values, resulting in the  final detected node set
\begin{align*}
    \widehat{V} = \big\{(x_i, y_i)\big\}_{i=1}^{N} \subset \mathbb{R}^2.
\end{align*}

\textbf{3) Node descriptor extraction}. For each detected node $i$, the refined coordinates are mapped to the local feature space $(x_i^L, y_i^L) = \left( x_i \cdot W_L / W,\; y_i \cdot H_L / H \right)$. We extract a local node descriptor by bilinearly sampling $\mathbf{F}_{\text{local}}$ at $(x_i^L, y_i^L)$:
\begin{align*}
    \boldsymbol{l}_i = \text{GridSample}\big(\mathbf{F}_{\text{local}}, (x_i^L, y_i^L)\big) \in \mathbb{R}^{D_N},
\end{align*}
where $D_N \in \mathbb{Z}^+$ is the node descriptor dimension. These descriptors encode visual and geometric features around each node and provide node-level representations for edge reasoning.


\subsubsection{Edge Reasoning}
Having obtained the detected node set $\widehat{V}(\mathbf{I})$ and their local descriptors, we infer the adjacency matrix $\widehat{\mathbf{A}}(\mathbf{I})$ by predicting edges over the detected nodes.
Edge inference contains three stages: 1) global topology encoding, 2) sparse candidate edge generation, and 3) topology-aware edge reasoning.

\textbf{1) Global topology encoding} (block (\#3) of Fig.~\ref{fig:framework}). 
Edge connectivity in node-link images often depends on image-level structure, such as global layout and long-range connectivity. To incorporate such image-level context into edge reasoning, we introduce a transformer augmented with learnable topology tokens. Given the global feature map $\mathbf{F}_{\text{global}}\in\mathbb{R}^{H_G \times W_G \times D_G}$ (Eq.~\ref{eq:global}), we flatten it into a sequence of spatial tokens and prepend $K_T \in \mathbb{Z}^+ $ learnable topology tokens $\mathbf{Z}_{\text{topo}} \in \mathbb{R}^{K_T \times D_G}$. The resulting sequence is processed by a transformer parameterized by ${\boldsymbol{\theta}_{T}}$ 
\begin{align*}
    \mathbf{F}_{\text{concat}}=\text{Transformer}_{\boldsymbol{\theta}_{T}}
    \big[\mathbf{Z}_{\text{topo}} \,;\,\text{Flatten}(\mathbf{F}_{\text{global}}) \big],
\end{align*}
where $\mathbf{F}_{\text{concat}}\in \mathbb{R}^{(K_T + H_G W_G) \times D_G}$. Through self-attention, the topology tokens attend to all spatial tokens to aggregate image-level patterns, and subsequently propagate this global context back to every spatial token.
Since topology tokens serve only as image-level topology carriers and have no spatial correspondence, we discard them after this interaction and retain the topology-enhanced spatial tokens, which are reshaped back into the global feature map:
\begin{align*}
    \mathbf{F}_{\text{global}}^{\prime} = \text{Reshape}\big(\mathbf{F}_{\text{concat}}[K_T:]\big) \in \mathbb{R}^{H_G \times W_G \times D_G },
\end{align*}
The resulting feature map $\mathbf{F}_{\text{global}}^{\prime}$ encodes global structural context and is subsequently used to guide candidate edge selection and adjacency structure reasoning.

\textbf{2) Sparse candidate edge generation} (block (\#4) of Fig.~\ref{fig:framework}).
Existing approaches typically produce candidate edges using fixed KNN-based methods \cite{zhang2021challenges, wang2024query, wang2025data}, lacking adaptability to diverse node-link images.  
To address this limitation, we introduce an attention-sparse neighbor sampler (\textbf{ASNS}) that learns to select a high-recall set of candidate node pairs.
For each detected node $i$, we first extract the global context
$\boldsymbol{g}_i \in \mathbb{R}^{D_G}$ by sampling
$\mathbf{F}_{\text{global}}^{\prime}$ at the node location. Let $N_h\in\mathbb{Z}^+$ be the number of heads and $d\in\mathbb{Z}^+$ the output dimension of each head, with $D_N = N_h \cdot d$.
We compute a compatibility score between node $i$ and $j$:
\begin{align*}
s_{ij}
=
\frac{1}{N_h}\sum_{m=1}^{N_h}
\frac{
\big(\mathbf{W}_{q}^{m} \boldsymbol{l}_i\big)^{\top}
\big(\mathbf{W}_{k}^{m} \boldsymbol{g}_j\big)
}{\sqrt{d}}
\in \mathbb{R},
\end{align*}
where $\mathbf{W}_{q}^{m}\in\mathbb{R}^{d \times D_N}$ and
$\mathbf{W}_{k}^{m}\in\mathbb{R}^{d \times D_G}$
are learnable projections for the $m$-th head, with all parameters collected as
$\boldsymbol{\theta}_{S}=\{\mathbf{W}_{q}^{m},\mathbf{W}_{k}^{m}\}_{m=1}^{N_h}$.
Collecting all scores into $\mathbf{S} \in \mathbb{R}^{N \times N}$, we apply the entmax activation \cite{peters2020sparse, correia2019adaptively}: 
\begin{align} 
\label{eq:asns}
\mathbf{P} = \operatorname{entmax}_{\alpha_{\text{ent}}}(\mathbf{S}) \in \mathbb{R}^{N \times N}, 
\end{align} 
yielding a sparse attention distribution over potential neighbors for all nodes. The sparsity parameter $\alpha_{\text{ent}}\in[1,2]$ controls the degree of sparsity in the attention distribution, where larger values generally produce sparser outputs. For each node $i$, we retain the top-$K \in \mathbb{Z}^+$ neighbors according to $\mathbf{P}_{i,:}$ to 
form a binary adjacency mask $\mathbf{M}_{\text{cand}} \in \{0,1\}^{N \times N}$. This learned mask focuses edge reasoning on a small set of plausible node pairs.

\textbf{3) Topology-aware edge reasoning} (block (\#5) in Fig.~\ref{fig:framework}). 
Given the candidate edge set defined by $\mathbf{M}_{\text{cand}}$, we infer the adjacency matrix by learning true edges from all candidates.
Rather than predicting each edge independently, we formulate edge reasoning as a structured problem, where the existence of an edge depends not only on visual evidence and geometric constraints, but also on interactions with other edges.
For each candidate node pair $(i,j)$, we construct an edge representation that integrates the following complementary features (3-a to 3-c) and jointly process all candidate edges using a line-graph transformer (3-d) that treats edges as tokens, enabling explicit modeling of edge dependencies.

3-a) \textit{Visual features.} 
To capture visual cues associated with a potential connection between nodes $i$ and $j$, we sample visual features along a smoothly curved path connecting their coordinates. 
Specifically, we model the path using a quadratic B\'ezier curve \cite{farin2014curves}:
\begin{align*}
\mathbf{c}_{ij}(t)
&=
(1-t)^2 (x_i,y_i)
+ 2t(1-t)
\left(
\frac{(x_i,y_i)+(x_j,y_j)}{2}
+\boldsymbol{\delta}_{ij}
\right) \\
&\quad
+ t^2 (x_j,y_j), \quad 
\text{and}\quad \boldsymbol{\delta}_{ij}=\text{MLP}_{\boldsymbol{\theta}_{\delta}}
([\boldsymbol{l}_i, \boldsymbol{l}_j]) \in \mathbb{R}^2,
\end{align*}
where $t \in [0,1]$ and $\boldsymbol{\delta}_{ij}$ is a learned control offset representing a deviation relative to the midpoint between nodes $i$ and $j$.
We sample $N_v \in \mathbb{Z}^+$ evenly spaced interpolation points 
\begin{align*}
    (x_{ij}^{n}, y_{ij}^{n}) = \mathbf{c}_{ij}\!\left(\frac{n}{N_v+1}\right),
    \quad n = 1,\dots,N_v.
\end{align*}
At each sampled location, we aggregate visual context within a $r_n \in \mathbb{Z}^+$-radius neighborhood to obtain descriptors $\{\mathbf{v}_{ij}^{n}\}_{n=1}^{N_v}$, which are then processed by a transformer parameterized by $\boldsymbol{\theta}_{\text{vis}}$
\begin{align}
\label{eq:visual}
    \boldsymbol{f}_{\text{vis}}
    = \mathrm{Transformer}_{\boldsymbol{\theta}_{\text{vis}}}\Big(\{\mathbf{v}_{ij}^{n}\}_{n=1}^{N_v}\Big)
    \in \mathbb{R}^{D_{\text{vis}}},
\end{align}
where $D_{\text{vis}} \in \mathbb{Z}^+$ denotes the visual feature dimension. 
This representation encodes stroke continuity and appearance consistency along the candidate connection.

3-b) \textit{Spatial features.} For each candidate edge $(i,j)$, we encode geometric constraints using three spatial features:

\noindent
i) the normalized Euclidean distance
\begin{align*}
d_{ij} = \frac{\|(x_j, y_j) - (x_i, y_i)\|_2}{\sqrt{H^2 + W^2}} \in [0,1].
\end{align*}
ii) the relative direction
\begin{align}
\label{eq:spatial}
\mathbf{u}_{ij} = \frac{(x_j - x_i, y_j - y_i)}{\|(x_j, y_j) - (x_i, y_i)\|_2} \in \mathbb{R}^2.
\end{align}
iii) a path consistency score measuring the minimum normalized distance from other nodes $k$ to the connection path $\mathbf{c}_{ij}(t)$:
\begin{align*}
    o_{ij}
=
\min\left\{
1,\;
\min_{k \notin \{i,j\}}
\frac{
\min_{t \in [0,1]}
\big\|
(x_k, y_k) - \mathbf{c}_{ij}(t)
\big\|_2
}{
\|(x_j, y_j) - (x_i, y_i)\|_2
}
\right\}
\in [0,1].
\end{align*}
A large score $o_{ij}$ indicates an unobstructed path, supporting the direct connectivity between $i$ and $j$.
The final spatial features are given by $\boldsymbol{f}_{\text{spatial}} = [d_{ij}, \mathbf{u}_{ij}, o_{ij}] \in \mathbb{R}^4$.

3-c) \textit{Topology features.} For each candidate edge $(i,j)$, we sample the topology-enhanced feature map $\mathbf{F}_{\text{global}}^\prime$ at the two endpoint locations to obtain global node descriptors
\begin{align}
\label{eq:topology}
    \boldsymbol{g}_i = \text{GridSample}\Big(\mathbf{F}_{\text{global}}^\prime, (x_i^G, y_i^G)\Big) \in \mathbb{R}^{D_G},
\end{align}
where $(x_i^G,y_i^G) = (x_i \cdot W_G/W,\; y_i \cdot H_G/H)$. $\boldsymbol{g}_j$ is defined analogously.

3-d) \textit{Line-graph transformer for adjacency prediction.} For each candidate edge $(i,j)$, we construct an edge-level representation $\boldsymbol{f}_{ij} \in \mathbb{R}^{D_E} $ with dimension $D_E \in\mathbb{Z}^+$ by combining
all the above features and positional encoding:
\begin{align*}
    \boldsymbol{f}_{ij} = \text{MLP}_{\boldsymbol{\theta}_{1}}\big([\boldsymbol{f}_{\text{spatial}},\boldsymbol{f}_{\text{vis}},  \boldsymbol{g}_i, \boldsymbol{g}_j]\big) + \text{MLP}_{\boldsymbol{\theta}_{2}}\big([x_i, y_i, x_j, y_j]\big).
\end{align*}
All candidate edges are collected into an edge-token tensor
\begin{align*}
\mathbf{F}_{\text{edge}}
=
\Big\{\boldsymbol{f}_{ij}\mid \mathbf{M}_{\text{cand}}[i,j]=1\Big\}
\;\in\;
\mathbb{R}^{(NK)\times D_E}.
\end{align*}
We perform joint reasoning over these edge tokens by applying a line-graph transformer ($\textbf{LineGT}$), where attention is restricted to incident edges that share a common endpoint \cite{cai2021line}. This enables capturing structured interactions among adjacent edges, such as degree patterns, while avoiding interactions between unrelated pairs. Since ASNS retains at most $K$ candidate neighbors for each of the $N$ detected nodes, LineGT only needs to model interactions among a bounded number of incident candidate edges per node, resulting in complexity $O(NK^2)$. 
After joint processing by LineGT with parameters $\boldsymbol{\theta}_{\text{line}}$, each edge token is mapped to a scalar logit:
\begin{align}
\label{eq:line_GT}
    \mathbf{E}_{\text{logits}} = \text{MLP}_{\boldsymbol{\theta}_{3}}\Big(\text{LineGT}_{\boldsymbol{\theta}_{\text{line}}}\big(\mathbf{F}_{\text{edge}}\big)\Big) \in \mathbb{R}^{N \times K}.
\end{align}
These logits are written back to a matrix $\mathbf{A}_{\text{logits}}\in\mathbb{R}^{N\times N} $ at the candidate locations indicated by $\mathbf{M}_{\text{cand}}$, while non-candidate entries 
are masked out and kept as non-edges.
We then enforce symmetry and obtain the edge probability matrix via a sigmoid activation 
\begin{align*}
    \mathbf{A}_{\text{pred}} = \sigma\Big(\frac{\mathbf{A}_{\text{logits}} + \mathbf{A}_{\text{logits}}^\top}{2}\Big) \in [0,1]^{N \times N}.
\end{align*}
The final predicted adjacency matrix $\widehat{\mathbf{A}}$ is obtained by thresholding $\mathbf{A}_{\text{pred}}$ with an edge threshold.


\subsection{Training}
Training VisAdj involves two key objectives:
i) establishing consistent correspondences between predicted and ground-truth nodes, and
ii) jointly optimizing node detection, sparse candidate edge generation, and final edge reasoning
within a unified objective.
In the following, we describe the training procedure in detail.


\textbf{Dataset preparation.}
To stabilize the early optimization of edge reasoning, we adopt a teacher-forcing strategy~\cite{bengio2015scheduled, lamb2016professor}, where edge prediction during training is conditioned on ground-truth nodes rather than detected nodes.
This prevents early-stage node detection errors, including node omissions and inaccurate localization, from corrupting candidate generation and edge supervision.
However, using ground-truth nodes throughout training may introduce a train-test discrepancy, since edge reasoning is conditioned on predicted nodes at inference time.
To mitigate this discrepancy, we adopt a scheduled node-conditioning strategy that gradually exposes the edge reasoning module to predicted nodes during training.
Let 
$V^{\mathrm{gt}}=\{(x_j^{\mathrm{gt}},y_j^{\mathrm{gt}})\}_{j=1}^{N_{\mathrm{gt}}}$ 
denote the ground-truth node set and 
$\widehat{V}=\{(x_i,y_i)\}_{i=1}^{N}$ 
denote the predicted node set.
When the teacher-forcing strategy is used, we first perturb the ground-truth node coordinates by adding Gaussian noise:
\begin{align*}
\widetilde{V}^{\mathrm{gt}}
=
\left\{
(x_j^{\mathrm{gt}},y_j^{\mathrm{gt}})+\boldsymbol{\epsilon}_j
\right\}_{j=1}^{N_{\mathrm{gt}}},
\qquad
\boldsymbol{\epsilon}_j \sim \mathcal{N}(\mathbf{0},\sigma_g^2\mathbf{I}),
\end{align*}
where $\sigma_g\in \mathbb{R}^+$ is the noise standard deviation and $N_{\mathrm{gt}}\in\mathbb{Z}^+$ denotes the number of ground-truth nodes. 
At each training epoch $e$, the node set used for edge reasoning, which we denote as
$V_{\mathrm{train}}=\{(x_i^{\mathrm{train}},y_i^{\mathrm{train}})\}_{i=1}^{N_{\mathrm{train}}}$, is sampled by
\begin{align*}
V_{\mathrm{train}}
=
\begin{cases}
\widetilde{V}^{\mathrm{gt}}, & \text{w.p.}\,\,\, p_e,\\
\widehat{V}, & \text{w.p.}\,\,\, 1-p_e,
\end{cases}
\quad \text{and} \quad
p_e = \max\left(0, 1-\frac{e}{T_s}\right),
\end{align*}
where $p_e \in [0, 1]$ represents the teacher-forcing probability and $T_s\in\mathbb{Z}^+$ denotes the number of epochs over which $p_e$ decays to zero.
This schedule preserves stable edge supervision in early training and progressively shifts the conditioning distribution toward the inference-time setting, thereby reducing the train-test discrepancy.
At inference time, ground-truth node coordinates are unavailable; VisAdj first detects the node set and then performs edge reasoning based only on the predicted node coordinates.

\textbf{Node Matching.}
Since the training node set used for edge reasoning may be unordered and may differ in cardinality from the ground truth, we establish correspondences based on spatial proximity using the Hungarian algorithm~\cite{cao2016querying}.
Specifically, we compute a minimum-cost bipartite matching between $V_{\mathrm{train}}$ and $V^{\mathrm{gt}}$ via
\begin{align}
\label{eq:node_match}
&\min_{\mathbf{X}}
\sum_{i=1}^{N_{\mathrm{train}}}\sum_{j=1}^{N_{\mathrm{gt}}}
\mathbf{X}_{ij}
\left\|
(x_i^{\mathrm{train}},y_i^{\mathrm{train}})
-
(x_j^{\mathrm{gt}},y_j^{\mathrm{gt}})
\right\|_2,
\\
\nonumber
\text{s.t. }\;
\sum_{j=1}^{N_{\mathrm{gt}}} \mathbf{X}_{ij}& \le 1,\quad
\sum_{i=1}^{N_{\mathrm{train}}} \mathbf{X}_{ij} \le 1,
\quad
\sum_{i=1}^{N_{\mathrm{train}}}\sum_{j=1}^{N_{\mathrm{gt}}}\mathbf{X}_{ij}=\min(N_{\mathrm{train}},N_{\mathrm{gt}})
\end{align}
where $\mathbf{X}\in \{0,1\}^{N_{\mathrm{train}} \times N_{\mathrm{gt}}}$ denotes the binary matching matrix.
After matching, only pairs whose distance is below the threshold $\tau_d\in\mathbb{R}^+$ are retained as valid correspondences.
The resulting correspondences are used to construct ground-truth adjacency labels in the training node space, providing supervision for both candidate generation and edge prediction.

\textbf{Loss Functions.} 
The training objective consists of three complementary losses to supervise
three stages of the pipeline: 
node prediction, candidate generation, and edge reasoning.

\textbf{1) Node mask loss.} 
Node detection is supervised by minimizing the difference between the predicted node likelihood map
$\widehat{\mathbf{M}}$ (Eq.~\ref{eq:node_mask}) and the ground-truth $\mathbf{M}$
\begin{align*}
\mathcal{L}_{\text{node}}\big(\boldsymbol{\theta}_{E}, \boldsymbol{\theta}_{L}, \boldsymbol{\theta}_{N}\big)
=
\lambda_{\text{ce}}\,
\ell_{\text{ce}}\big(\widehat{\mathbf{M}}, \mathbf{M}\big)
+
\lambda_{\text{mse}}\,
\ell_{\text{mse}}\big(\widehat{\mathbf{M}}, \mathbf{M}\big),
\end{align*}
where $\lambda_{\text{ce}}, \lambda_{\text{mse}} \in \mathbb{R}^+$ are balancing weights. $\ell_{\text{ce}}$ denotes a weighted pixel-wise cross-entropy loss addressing foreground-background imbalance,
and $\ell_{\text{mse}}$ denotes the mean squared error measuring pixel-wise deviations
between $\widehat{\mathbf{M}}$ and $\mathbf{M}$.

\textbf{2) Coverage Loss.} The attention-sparse neighbor sampler outputs a sparse attention distribution $\mathbf{P}_{i,:}$ (Eq. \ref{eq:asns}) over potential neighbors for each node. To encourage high-recall candidate generation, we supervise this distribution using a smoothed target derived from the ground-truth adjacency.
For each matched training node $i$, the target distribution $\mathbf{T}_{i} \in \mathbb{R}^{N_{\mathrm{train}}}$ is defined as
\begin{align*}
    \mathbf{T}_{ij}
=
(1-\alpha_s)\,
\frac{\mathbb{I}[j\in\mathcal{N}_{\text{gt}}(i)]}{|\mathcal{N}_{\text{gt}}(i)|}
+
\alpha_s\,\frac{1}{N_{\mathrm{train}}-1}, \quad j\neq i,
\end{align*}
where $\mathcal{N}_{\mathrm{gt}}(i)$ denotes the ground-truth neighbor set mapped to the training node space via Eq.~\ref{eq:node_match}, $\mathbb{I}$ is the indicator function, and $\alpha_s \in [0,1]$ is a smoothing factor. The coverage loss is defined as the cross-entropy between predicted and target neighbor distributions:
\begin{align*}
    \mathcal{L}_{\text{cover}}\big(\boldsymbol{\theta}_{G}, \boldsymbol{\theta}_{T}, \boldsymbol{\theta}_{S}\big)
=
-\frac{1}{N_{\mathrm{train}}}
\sum_{i=1}^{N_{\mathrm{train}}}
\sum_{j=1,\,j\neq i}^{N_{\mathrm{train}}}
\mathbf{T}_{ij} \log\big(\mathbf{P}_{ij}\big).
\end{align*}
This loss encourages ASNS to assign non-negligible attention mass to all ground-truth neighbors of each node, thereby promoting high-recall candidate selection.

\textbf{3) Edge loss.} Edge reasoning is supervised by minimizing a focal loss defined on node-matched candidate adjacency entries.
Let $\mathcal{C}=\{(i,j)\mid \mathbf{M}_{\mathrm{cand}}[i,j]=1\}$ denote the candidate edge set. For each candidate pair $(i,j)\in\mathcal{C}$, we define
$p_{ij}=\mathbf{A}_{\mathrm{pred}}[i,j]$ if $(i,j)$ is a ground-truth edge after node matching, and  $p_{ij}=1-\mathbf{A}_{\mathrm{pred}}[i,j]$ otherwise. The edge loss is then defined as
\begin{align*}
\mathcal{L}_{\mathrm{edge}}\big(\boldsymbol{\theta}_{\delta}, \boldsymbol{\theta}_{\text{vis}}, \boldsymbol{\theta}_{1}, \boldsymbol{\theta}_{2}, \boldsymbol{\theta}_{3}, \boldsymbol{\theta}_{\text{line}}\big)
=
-\frac{1}{|\mathcal{C}|}
\sum_{(i,j)\in\mathcal{C}}
\alpha_{ij}(1-p_{ij})^\gamma \log(p_{ij})
\end{align*}
where $\alpha_{ij}$ is set to $\alpha_f$ for positive candidate edges and $1-\alpha_f$ for negative candidate edges. $\alpha_f \in[0,1]$ and $\gamma\in\mathbb{R}^+$ denote the balance and focusing parameters of focal loss \cite{lin2017focal}, respectively. 
This formulation mitigates the severe class imbalance in graphs by
down-weighting easy negative pairs and emphasizing hard edge decisions.
Although the loss is applied per adjacency entry, edge probabilities are predicted jointly by the line-graph transformer, enabling structured dependencies among incident edges to be learned.

\textbf{Total loss.} 
The overall objective of VisAdj is given by
\begin{align*}
    \mathcal{L}_{\text{total}} \big(\boldsymbol{\theta}_{E}, \boldsymbol{\theta}_{L}, \boldsymbol{\theta}_{N}, \boldsymbol{\theta}_{G}, \boldsymbol{\theta}_{T}, \boldsymbol{\theta}_{S}, \boldsymbol{\theta}_{\delta}, \boldsymbol{\theta}_{\text{vis}}, \boldsymbol{\theta}_{1}, \boldsymbol{\theta}_{2}, \boldsymbol{\theta}_{3}, \boldsymbol{\theta}_{\text{line}} \big) \\ = \lambda_{\text{node}} \cdot \mathcal{L}_{\text{node}}  + \lambda_{\text{cover}} \cdot \mathcal{L}_{\text{cover}} + \lambda_{\text{edge}} \cdot \mathcal{L}_{\text{edge}},
\end{align*}
where $\lambda_{\text{node}}, \lambda_{\text{cover}}, \lambda_{\text{edge}} \in\mathbb{R}^+$ are loss weights.

\textbf{Optimization scheme.} 
The vision encoder is fine-tuned using Low-Rank Adaptation (LoRA)~\cite{hu2022lora},
while all remaining modules are trained from scratch.
All trainable components are optimized jointly using the total loss
$\mathcal{L}_{\text{total}}$.

\begin{table}[t]
\caption{Dataset statistics.}
\label{tab:dataset}
\centering
\setlength{\tabcolsep}{4pt}
\begin{tabular}{lccccc}
\toprule
Dataset & Category & Image & \#Train & \#Val & \#Test \\
\midrule
Synthetic & Graph & RGB & 2763 & 395 & 788 \\
Toulouse & Road & Binary & 80357 & 11679 & 18998 \\
US-Cities & Road & RGB & 32740 & 1931 & 5730 \\
OCTA500 & Vessel & Grayscale & 11461 & 753 & 1505 \\
\bottomrule
\end{tabular}
\end{table}

\section{Experiments}

\subsection{Experimental Settings}

\subsubsection{Dataset}
We evaluate VisAdj on four datasets spanning synthetic graphs, road and vessel networks. These datasets cover various graph types and visual styles.
The detailed dataset statistics are summarized in Table~\ref{tab:dataset} and each dataset is described below.

\noindent
\textbf{1) Synthetic Dataset}: We construct a synthetic node-link image dataset to evaluate VisAdj under diverse graph structures and visual layouts. The underlying graphs are collected from the House of Graphs \cite{coolsaet2023house}, covering planar/non-planar and tree/non-tree graphs. We generate graph layouts using the Fruchterman-Reingold and Kamada-Kawai algorithms~\cite{fruchterman1991graph,kamada1989algorithm}, and render them as RGB images at a resolution of $512\times512$. This dataset contains challenging structures such as frequent edge crossings and long-range connections, enabling evaluation beyond the regular patterns commonly observed in real-world networks.


\noindent
\textbf{2) Toulouse Dataset} \cite{belli2019image}: 
The Toulouse dataset consists of binary road network images extracted from map imagery, with a resolution of $64\times64$. This dataset features a relatively clean visual appearance and mostly local connectivity, serving as a real-world benchmark for evaluating performance on well-structured sparse planar graphs.

\noindent
\textbf{3) US-Cities Dataset}~\cite{he2020sat2graph}: 
US-Cities contains high-resolution satellite imagery from 20 U.S. cities. 
Following prior work~\cite{shit2022relationformer}, we extract overlapping $128\times128$ patches from the original 180 tiles of resolution $2048\times2048$ to construct node-link images. Compared with Toulouse, this dataset contains more cluttered backgrounds, occlusions, and complex junctions, enabling more challenging evaluation on realistic large-scale road networks. 


\noindent
\textbf{4) OCTA500 Dataset}~\cite{li2024octa}: 
OCTA500 is a retinal vessel imaging dataset that contains optical coherence tomography angiography scans with the corresponding vessel segmentation maps. 
We extract overlapping $256\times256$ patches from 500 scans with $6\,\mathrm{mm}\times6\,\mathrm{mm}$ and $3\,\mathrm{mm}\times3\,\mathrm{mm}$ fields of view to increase data diversity. 
This dataset is characterized by tree-like vessel graphs with fine-scale branches and low-contrast structures, making it suitable for evaluating graph recovery under noisy medical imaging conditions.


\begin{table*}[t]
\caption{Main comparison results across all datasets. The best performance for each metric is highlighted in \textbf{bold}.}
\label{tab:main_results}
\centering
\renewcommand{\arraystretch}{1.1}
\begin{tabular}{l @{\hskip 6pt} c @{\hskip 8pt} c @{\hskip 6pt}c@{\hskip 8pt} c@{\hskip 6pt} c @{\hskip 6pt}c@{\hskip 8pt} c@{\hskip 6pt} c}
\toprule
&& \multicolumn{2}{c}{Graph-level Metrics} 
& \multicolumn{3}{c}{Subgraph-level Metrics (TOPO)} 
& \multicolumn{2}{c}{Element-level Metrics}  \\
\cmidrule(lr){3-4} \cmidrule(lr){5-7} \cmidrule(lr){8-9} 
Dataset & Method 
& GIR$\uparrow$ (\%) 
& GED$\downarrow$ 
& Precision$\uparrow$ (\%) 
& Recall$\uparrow$ (\%) 
& F1$\uparrow$ (\%) 
& Edge-F1$\uparrow$ (\%) 
& Node-F1$\uparrow$ (\%) \\
\midrule
\multirow{5}{*}{Synthetic}
& G-SAM-Road++   & \meanstd{43.17}{0.73}  & \meanstd{18.54}{2.24} & \meanstd{56.79}{1.36} & \meanstd{55.34}{1.08} & \meanstd{56.06}{1.18} & \meanstd{53.25}{1.24} & \meanstd{92.23}{1.37} \\
& G-RNGDet++     & \meanstd{26.31}{1.53} & \meanstd{19.81}{2.36} & \meanstd{55.73}{1.31} & \meanstd{45.85}{1.86} & \meanstd{50.31}{1.67} & \meanstd{44.64}{1.61} & \meanstd{88.82}{1.58} \\
& Any2Graph     & \meanstd{53.79}{0.57} & \meanstd{17.96}{1.67} & \meanstd{57.35}{1.68} & \meanstd{55.10}{1.43} & \meanstd{56.20}{1.40} & \meanstd{59.14}{1.58} & \meanstd{92.69}{1.24} \\
& Sat2Graph     & \meanstd{20.91}{1.93} & \meanstd{21.35}{1.85} & \meanstd{51.42}{1.46} & \meanstd{47.05}{1.57} & \meanstd{49.14}{1.63} & \meanstd{40.35}{1.59} & \meanstd{88.46}{1.67} \\
& VisAdj         & \textbf{\meanstd{73.02}{0.46}} & \textbf{\meanstd{0.50}{0.42}} & \textbf{\meanstd{98.78}{0.41}} & \textbf{\meanstd{98.67}{0.49}} & \textbf{\meanstd{98.72}{0.43}} & \textbf{\meanstd{95.09}{1.25}} & \textbf{\meanstd{99.23}{0.37}} \\
\midrule
\multirow{5}{*}{Toulouse}
& G-SAM-Road++    & \meanstd{86.75}{0.51} & \meanstd{0.60}{0.14} & \meanstd{98.19}{0.35} & \meanstd{95.70}{0.46} & \meanstd{96.93}{0.41} & \meanstd{96.35}{0.32} & \meanstd{98.89}{0.29} \\
& G-RNGDet++     & \meanstd{77.66}{1.45} & \meanstd{0.75}{0.50} & \meanstd{95.12}{0.83} & \meanstd{94.52}{0.52} & \meanstd{94.82}{0.36} & \meanstd{91.24}{0.76} & \meanstd{98.67}{0.14} \\
& Any2Graph     & \meanstd{93.45}{0.21} & \meanstd{0.13}{0.01} & \meanstd{98.26}{0.36} & \meanstd{97.53}{0.28} & \meanstd{97.89}{0.33} & \meanstd{98.87}{0.59} & \meanstd{99.31}{0.07} \\
& Sat2Graph     & \meanstd{69.86}{1.73} & \meanstd{1.35}{0.29} & \meanstd{92.55}{0.77} & \meanstd{91.59}{0.35} & \meanstd{92.07}{0.58} & \meanstd{87.63}{0.79} & \meanstd{94.86}{0.56} \\
& VisAdj           & \textbf{\meanstd{94.38}{0.33}} & \textbf{\meanstd{0.11}{0.02}} & \textbf{\meanstd{98.33}{0.43}} & \textbf{\meanstd{98.15}{0.45}} & \textbf{\meanstd{98.24}{0.41}} & \textbf{\meanstd{99.15}{0.49}} & \textbf{\meanstd{99.60}{0.11}} \\
\midrule
\multirow{5}{*}{US-Cities}
& G-SAM-Road++   & \meanstd{58.31}{1.13} & \meanstd{3.45}{0.31} & \meanstd{87.55}{0.82} & \meanstd{86.95}{0.86} & \meanstd{87.25}{0.85} & \meanstd{79.77}{1.17} & \meanstd{92.53}{0.90} \\
& G-RNGDet++     & \meanstd{48.74}{1.64} & \meanstd{5.69}{0.49} & \meanstd{79.63}{1.28} & \meanstd{80.03}{1.14} & \meanstd{79.83}{1.21} & \meanstd{72.04}{1.14} & \meanstd{90.20}{1.21} \\
& Any2Graph     & \meanstd{55.64}{0.74} & \meanstd{3.86}{0.58} & \meanstd{86.07}{0.83} & \meanstd{85.38}{1.27} & \meanstd{85.72}{1.10} & \meanstd{76.94}{1.10} & \meanstd{91.84}{0.92} \\
& Sat2Graph     & \meanstd{32.48}{1.13} & \meanstd{6.23}{0.23} & \meanstd{61.38}{0.65} & \meanstd{62.45}{0.84} & \meanstd{61.91}{0.73} & \meanstd{54.20}{0.43} & \meanstd{89.45}{0.95} \\
& VisAdj          & \textbf{\meanstd{68.57}{0.74}} & \textbf{\meanstd{0.57}{0.75}} & \textbf{\meanstd{93.86}{0.73}} & \textbf{\meanstd{93.06}{1.42}} & \textbf{\meanstd{93.46}{1.37}} & \textbf{\meanstd{88.54}{1.32}} & \textbf{\meanstd{96.68}{0.66}} \\
\midrule
\multirow{5}{*}{OCTA500}
& G-SAM-Road++   & \meanstd{51.76}{0.93} & \meanstd{3.12}{0.39} & \meanstd{81.91}{1.32} & \meanstd{82.76}{1.03} & \meanstd{82.32}{1.14} & \meanstd{77.66}{1.37} & \meanstd{91.25}{0.91} \\
& G-RNGDet++     & \meanstd{35.27}{1.48} & \meanstd{4.55}{0.53} & \meanstd{69.80}{1.48} & \meanstd{66.45}{1.18} & \meanstd{68.08}{1.31} & \meanstd{66.57}{1.46} & \meanstd{89.08}{1.11} \\
& Any2Graph     & \meanstd{45.79}{1.27} & \meanstd{3.57}{0.17} & \meanstd{79.05}{1.17} & \meanstd{77.26}{1.56} & \meanstd{78.14}{1.32} & \meanstd{74.95}{1.72} & \meanstd{90.46}{0.92} \\
& Sat2Graph     & \meanstd{27.15}{0.83} & \meanstd{6.54}{0.98} & \meanstd{56.42}{0.96} & \meanstd{55.84}{0.75} & \meanstd{56.13}{0.83} & \meanstd{49.09}{0.99} & \meanstd{86.28}{0.78} \\
& VisAdj      & \textbf{\meanstd{64.98}{0.59}} & \textbf{\meanstd{0.63}{0.50}} & \textbf{\meanstd{88.85}{0.75}} & \textbf{\meanstd{91.30}{0.64}} & \textbf{\meanstd{90.06}{0.72}} & \textbf{\meanstd{90.27}{0.81}} & \textbf{\meanstd{96.33}{0.64}} \\
\bottomrule
\end{tabular}
\end{table*}



\subsubsection{Baselines}
We compare VisAdj with four typical graph inference baselines, including
\textbf{G-SAM-Road++} \cite{yin2025towards},
\textbf{G-RNGDet++} \cite{xu2023rngdet++},
\textbf{Any2Graph} \cite{krzakala2024any2graph}, and
\textbf{Sat2Graph} \cite{he2020sat2graph}.
Here, G-SAM-Road++ and G-RNGDet++ denote the graph reasoning modules in SAM-Road++~\cite{yin2025towards} and RNGDet++~\cite{xu2023rngdet++}, respectively.
These baselines cover complementary graph inference paradigms, including rule-based graph construction from visual predictions, sequential graph growing, and general end-to-end graph prediction.
All baselines are implemented using their official codebases and adapted to the image-to-adjacency-matrix setting.

\subsubsection{Evaluation Metrics}
We utilize the following metrics.

\textbf{1) Graph-level metrics.}
We report Graph Isomorphism Rate (\textbf{GIR})~\cite{hopcroft1974linear}, which measures the percentage of predicted graphs that are exactly isomorphic to the ground-truth graphs. 
GIR is a strict metric and only counts a prediction as correct when the entire graph structure is perfectly reconstructed.
We also report Graph Edit Distance (\textbf{GED})~\cite{gao2010survey}, which quantifies global structural similarity via the minimum number of node and edge edit operations required to transform the predicted graph into the ground-truth graph.
GIR and GED are computed using standard NetworkX implementations \cite{hagberg2007exploring}; for GED, we use a 300-second timeout per sample and report the mean over all test samples.

\textbf{2) Subgraph-level metrics.}
We adopt TOPO Precision, Recall, and F1 score~\cite{he2020sat2graph} to evaluate local topological consistency.
For each matched node, TOPO compares the edge sets of the induced $k$-hop neighborhoods in the prediction and ground truth, where we set $k=2$ in all experiments.
Compared with exact graph-level metrics, TOPO metrics are more robust to small node or edge errors, and better reflect local structural correctness.

\textbf{3) Element-level metrics.}
We report Node-F1 and Edge-F1 scores.
Node-F1 evaluates node detection accuracy after spatial matching, where a predicted node is considered correct if it is matched to a ground-truth node within a distance threshold of $4$ pixels for Toulouse and $8$ pixels for other datasets.
Edge-F1 measures pairwise connectivity accuracy over matched node pairs, isolating the quality of edge prediction from global graph-level correctness.


\subsubsection{Training Settings}
To ensure a fair comparison, VisAdj and baselines use the same vision backbone, namely the SAM-ViT-B encoder.
All models are trained for up to 200 epochs with early stopping based on validation performance.
Each experiment is repeated three times with different random seeds, and we report the mean and standard deviation of all evaluation metrics.
We optimize all trainable modules using AdamW~\cite{loshchilovdecoupled}.
The initial learning rate is set to $1\times10^{-3}$ for newly initialized modules and $1\times10^{-4}$ for LoRA fine-tuning of the image encoder backbone.
The learning rate follows a cosine decay schedule with a linear warmup of 10 epochs.
The batch size is set to 96 and the training is conducted using Distributed Data Parallel (DDP)~\cite{li2020pytorch} on 4 NVIDIA L40S GPUs.
For VisAdj, the key hyperparameters are set as follows. The entmax sparsity parameter is set to $\alpha_{\mathrm{ent}}=1.5$. The model uses $K_T=16$ topology tokens and a teacher-forcing decay length of $T_s=30$. For candidate generation, VisAdj retains $K=8$ neighbors on the Toulouse dataset and $K=12$ neighbors on the other datasets. Detailed hyperparameter configuration can be found in our public~\href{https://github.com/Jiahao-Xie-86/VisAdj}{GitHub~repository}.



\subsection{Main Results and Analysis}
This section analyzes the comparative results and also presents sanity checks
to validate that VisAdj works as designed.

\textbf{VisAdj consistently outperforms all baselines.} As shown in Table \ref{tab:main_results}, VisAdj achieves the best results across all four datasets and all evaluation metrics. On the Synthetic dataset, VisAdj improves the GIR from Any2Graph's $53.79\%$ to $73.02\%$, indicating a substantial improvement in recovering global graph structure. 
Similar gains are also observed on real-world data. For example, VisAdj achieves GIR scores of $68.57\%$ on US-Cities and $64.98\%$ on OCTA500, exceeding the strongest baseline, G-SAM-Road++, by clear margins.
Beyond exact graph matching, VisAdj also achieves stronger local topological consistency, obtaining the highest TOPO-F1 scores on all datasets. For instance, VisAdj increases TOPO-F1 from G-SAM-Road++'s $82.32\%$ to $90.06\%$ on OCTA500.  At the element level, VisAdj consistently delivers higher Edge-F1 scores, with Node-F1 exceeding $96\%$ across all datasets.
These results collectively confirm that VisAdj recovers both global and local graph structure more faithfully than existing methods across various domains.
In addition, the observed improvements are also substantially larger than the reported standard deviations. For instance, on the Synthetic dataset, VisAdj improves GIR by $19.23\%$ over the strongest baseline, while the corresponding standard deviations are below $0.6\%$. Similar margins are also observed on US-Cities and OCTA500, supporting that the improvements are robust across random seeds.

\textbf{Performance gains mainly come from improved edge reasoning.} According to Table \ref{tab:main_results}, Node-F1 scores are already high across all methods and datasets, with relatively small variance between VisAdj and the baselines. In contrast, Edge-F1 exhibits substantially larger gaps. For example, VisAdj improves Edge-F1 from Any2Graph's $59.14\%$ to $95.09\%$ on the Synthetic dataset, and from G-SAM-Road++'s $77.66\%$ to $90.27\%$ on the OCTA500 dataset.
These results demonstrate that VisAdj achieves significantly stronger edge reasoning capability than existing methods, directly validating the core idea and design of our model. 
By incorporating the global structural context and jointly reasoning over candidate edges with a line-graph transformer, VisAdj produces more discriminative and structurally consistent edge predictions, which directly translates into improved graph-level correctness.

\begin{table}[t]
    \caption{Performance of VisAdj with different image encoders on the Synthetic dataset.}
    \label{tab:encoder}
\renewcommand{\arraystretch}{1.1}
    \centering
    \begin{tabular}{lccc}
    \toprule
        Image encoder & Size (M) &  GIR$\uparrow$ (\%) & TOPO-F1$\uparrow$ (\%) \\
        \midrule
        SAM-ViT-B \cite{kirillov2023segment} & 91 & \meanstd{73.02}{0.46} & \meanstd{98.72}{0.43}   \\
        SAM-ViT-L \cite{kirillov2023segment} & 308& \meanstd{75.38}{0.39}  & \meanstd{98.76}{0.42} \\
        SAM-ViT-H \cite{kirillov2023segment} & 636&  \meanstd{76.73}{0.41} & \meanstd{98.81}{0.43}  \\
        SAM2-Hiera-S \cite{ravi2025sam} & 46 & \meanstd{74.87}{0.38} & \meanstd{98.73}{0.41}   \\
        SAM2-Hiera-B \cite{ravi2025sam} & 80.8 &  \meanstd{75.56}{0.35} & \meanstd{98.78}{0.39}   \\
        SAM2-Hiera-L \cite{ravi2025sam} & 224.4& \meanstd{78.08}{0.32} & \meanstd{98.84}{0.38}  \\
        SAM3 \cite{carion2025sam} & 900& \meanstd{78.23}{0.27} & \meanstd{98.89}{0.35}  \\
    \bottomrule
    \end{tabular}
\end{table}

\begin{figure*}[t]
    \centering
    \includegraphics[width=1\linewidth]{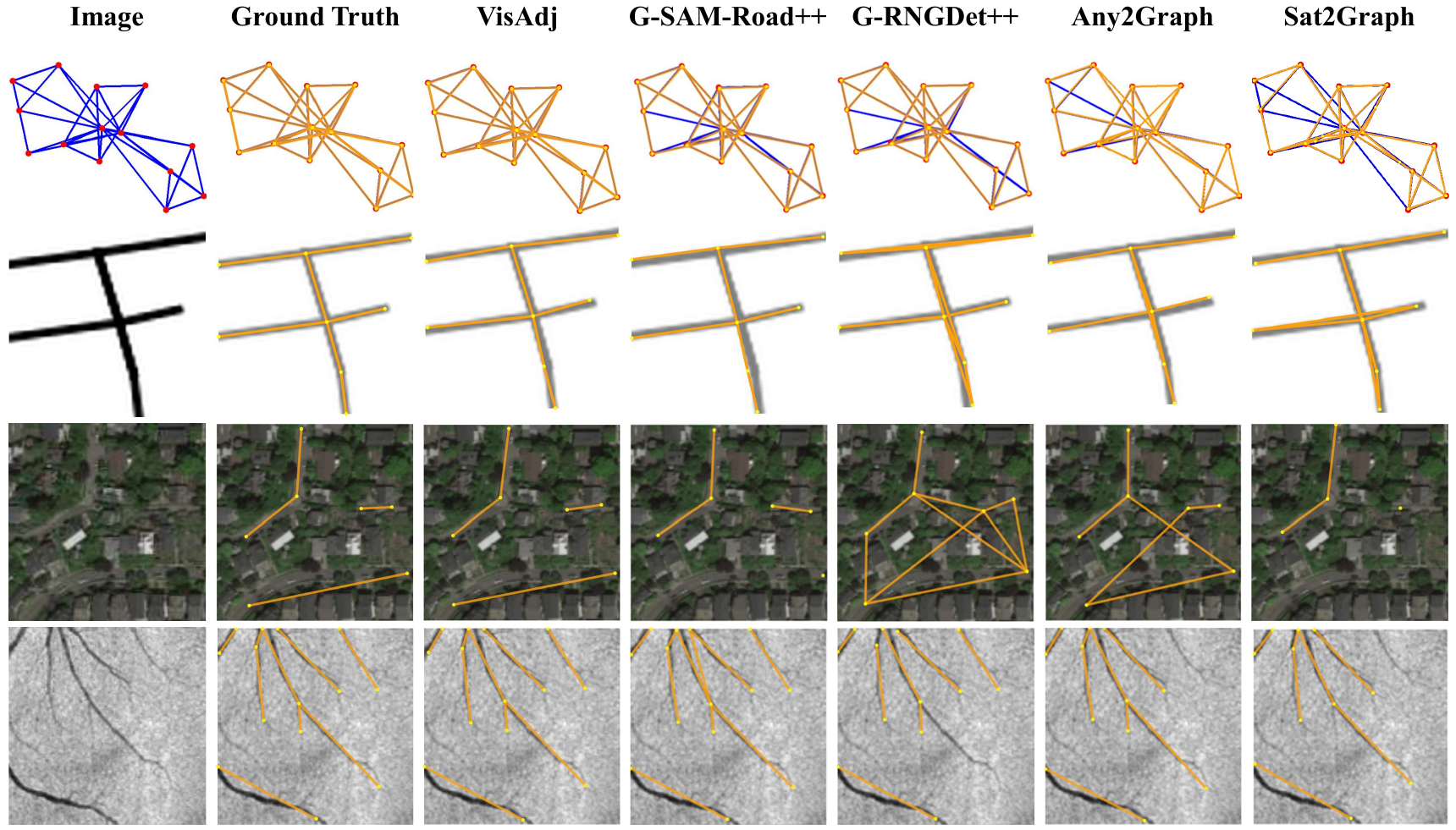}
    \caption{Qualitative results. Each row presents one example, with columns showing the input image, ground-truth graph, and predictions from VisAdj, G-SAM-Road++, G-RNGDet++, Any2Graph, and Sat2Graph, respectively. Rows correspond to examples from the Synthetic, Toulouse, US-Cities, and OCTA500 datasets. Predicted nodes are shown as yellow dots, and predicted edges as orange line segments.}
    \Description{A qualitative comparison table with four rows, one for each dataset. Each row shows an input image, the ground-truth graph, and predicted graphs from VisAdj, G-SAM-Road++, G-RNGDet++, Any2Graph, and Sat2Graph. VisAdj predictions generally match the ground-truth graph more closely, while baselines show more missing or spurious edges.}
    \label{fig:visualization}
\end{figure*}

\begin{figure*}[t]
    \centering
    \includegraphics[width=1\linewidth]{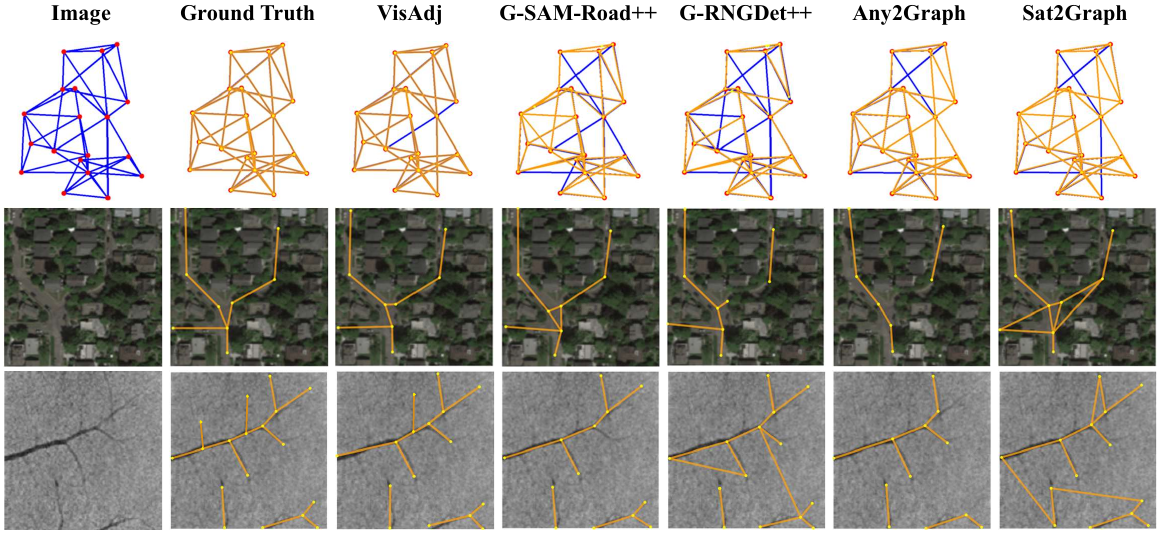}
    \caption{Representative failure cases. Each row presents one example, with columns showing the input image, ground-truth graph, and predictions from VisAdj, G-SAM-Road++, G-RNGDet++, Any2Graph, and Sat2Graph. Predicted nodes are shown as yellow dots, and predicted edges as orange line segments. Although VisAdj may still make incorrect adjacency decisions under dense crossings, cluttered backgrounds, or weak visual contrast, it generally produces better predictions than the baselines.}
    \Description{Qualitative examples of failure cases where VisAdj makes incorrect graph connections. The examples include dense crossings or cluttered visual regions where multiple edge configurations are plausible, causing missed edges or incorrect shortcut edges.}
    \label{fig:failure_cases}
\end{figure*}

\textbf{Stronger image encoders lead to better performance.}
Table~\ref{tab:encoder} shows that VisAdj can benefit from stronger image encoders. Within the SAM-ViT family, increasing model capacity leads to steady improvements in graph-level accuracy: GIR increases from $73.02\%$ with SAM-ViT-B to $76.73\%$ with SAM-ViT-H. This trend indicates that larger encoders provide richer visual features that better support graph structure reasoning.
Beyond model scale, replacing SAM with SAM2 further yields systematic gains. Under comparable sizes, SAM2-Hiera encoders consistently outperform their SAM-ViT counterparts. For example, GIR improves from $73.02\%$ using SAM-ViT-B to $75.56\%$ using SAM2-Hiera-B. These results echo the fact that SAM2 has stronger representation ability than SAM in general.
However, the performance gains diminish as the representation ability of the image encoder increases.
For example, replacing SAM2-Hiera-L with SAM3 only increases GIR from $78.08\%$ to $78.23\%$.
This suggests that once the image encoder offers sufficient representation capacity, visual encoding is no longer the primary bottleneck.
Further performance improvements therefore depend more on graph structure reasoning.

\textbf{VisAdj enables robust graph reasoning under visual ambiguity.}
Fig.~\ref{fig:visualization} qualitatively compares VisAdj with baselines. 
In visually challenging scenarios, such as cluttered backgrounds, closely spaced structures, and low-contrast vessel structures, the baselines often introduce spurious connections or miss true edges, indicating difficulty in inferring valid connectivity from local visual cues alone.
In contrast, VisAdj consistently recovers adjacency structures that more closely match the ground-truth. On the synthetic dataset, VisAdj preserves long-range connections while suppressing implausible crossings. On road network images, it accurately recovers junction connectivity and road continuity, even when road segments are closely spaced or partially occluded. On OCTA500 vessel images,
VisAdj correctly captures branching structures and vessel terminations while avoiding incorrect shortcuts.
We further analyze representative failure cases in Fig.~\ref{fig:failure_cases}.
VisAdj may still fail in densely connected regions with heavy edge crossings or visually ambiguous connections, where it may introduce incorrect shortcuts or miss weakly visible edges.
However, even in these challenging cases, VisAdj generally produces fewer errors than the baselines, with more accurate node detection and fewer incorrect or missing edges, indicating stronger robustness under complex visual conditions. These qualitative results underscore the effectiveness of VisAdj in adjacency reasoning across diverse visual domains. 

\begin{table}[t]
\caption{Inference runtime on the US-Cities dataset with a batch size of 1 under the same settings. Speedup is computed as the latency ratio between each baseline and VisAdj.}
\label{tab:runtime}
\renewcommand{\arraystretch}{1.1}
\centering
\begin{tabular}{lccc}
\toprule
Method & ms/image $\downarrow$ & images/s $\uparrow$ & Speedup $\uparrow$ \\
\midrule
G-SAM-Road++ & 88.18  & 11.34 & $1.40\times$ \\
G-RNGDet++   & 269.54 & 3.71  & $4.27\times$ \\
Any2Graph  & 102.25 & 9.78  & $1.62\times$ \\
Sat2Graph  & 147.93 & 6.76  & $2.34\times$ \\
\textbf{VisAdj} & \textbf{63.17} & \textbf{15.83} & \textbf{--} \\
\bottomrule
\end{tabular}
\end{table}

\textbf{Runtime efficiency.}
We further evaluate inference efficiency on the US-Cities dataset with batch size 1 under the same evaluation settings.
As reported in Table~\ref{tab:runtime}, VisAdj achieves the lowest latency and highest throughput among all compared methods, requiring only $63.17$ ms per image and processing $15.83$ images per second.
Compared with G-SAM-Road++, Any2Graph, Sat2Graph, and G-RNGDet++, VisAdj is approximately $1.40\times$, $1.62\times$, $2.34\times$, and $4.27\times$ faster, respectively.
These results show that the proposed graph reasoning method improves effectiveness without sacrificing inference efficiency.
The efficiency gain mainly comes from ASNS, which restricts edge prediction to a sparse set of plausible candidate pairs, and LineGT, which models dependencies only among incident candidate edges rather than performing dense all-pair interactions.

\begin{figure}
    \centering
    \includegraphics[width=1.0\linewidth]{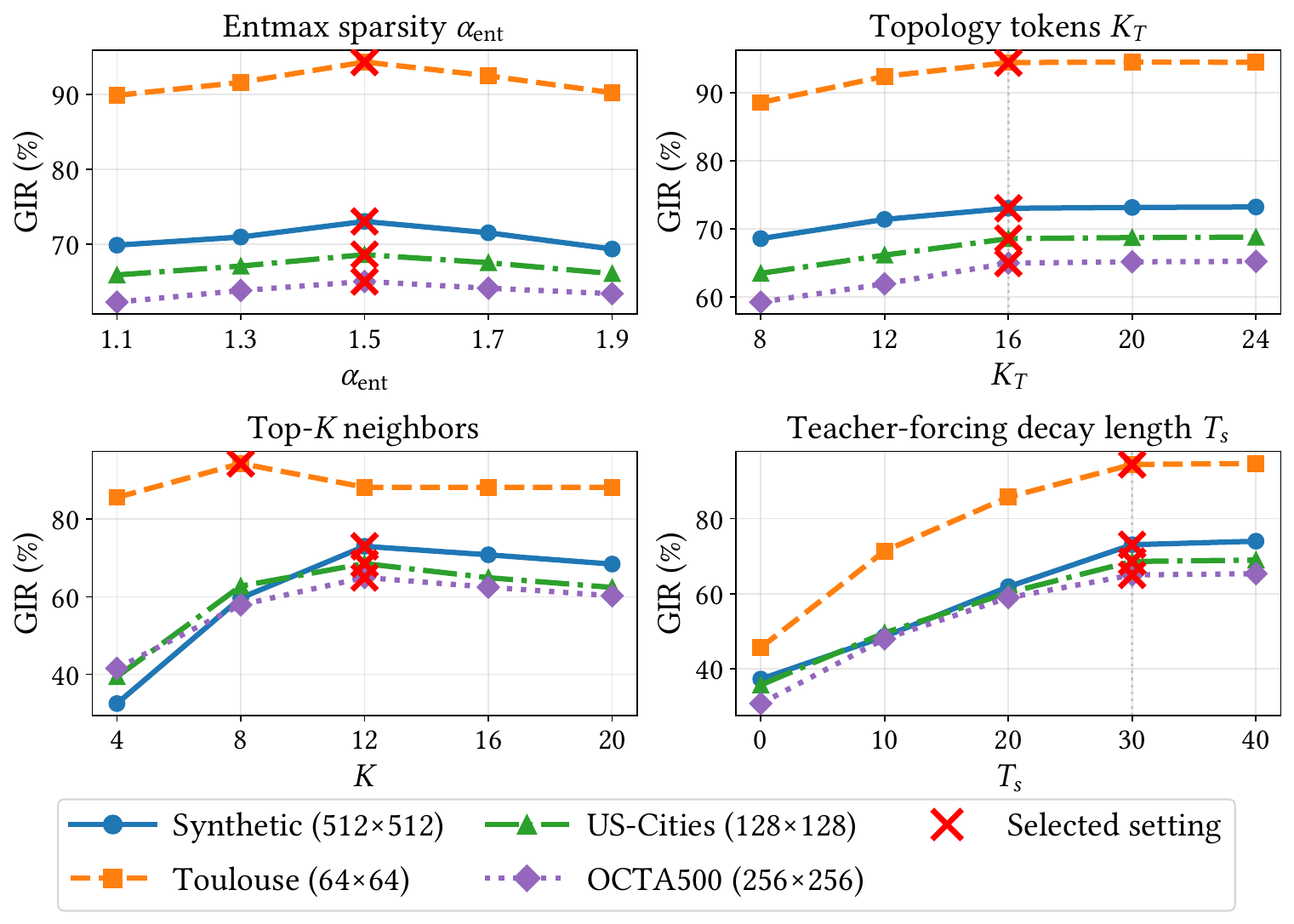}
    \caption{Sensitivity analysis of key hyperparameters on Synthetic, Toulouse, US-Cities, and OCTA500 datasets. }
    \Description{Sensitivity analysis of four key hyperparameters on different datasets.}
    \label{fig:sensitivity}
\end{figure}

\textbf{Analysis of sensitivity to key hyperparameters.}
We also analyze the sensitivity of VisAdj to four key hyperparameters: the entmax sparsity parameter $\alpha_{\text{ent}}$, the number of topology tokens $K_T$, the number of retained candidate neighbors $K$, and the teacher-forcing decay length $T_s$.
The results are summarized in Fig.~\ref{fig:sensitivity}. \textbf{1)} VisAdj is relatively robust to the sparsity parameter $\alpha_{\text{ent}}$, with $\alpha_{\text{ent}}=1.5$ generally achieving the best GIR across datasets.
This indicates that the model is not overly sensitive to the exact entmax sparsity level, as long as the attention distribution maintains a reasonable balance between sparsity and true-edge coverage.
\textbf{2)} For topology tokens, increasing $K_T$ to $16$ clearly improves performance, while further increasing $K_T$ only brings marginal gains. Since GIR becomes nearly saturated after $K_T=16$, we select $K_T=16$ to balance reconstruction accuracy and model size.
\textbf{3)} For Top-$K$ candidate selection, a small $K$ misses many true neighbors and leads to low edge recall, whereas a large $K$ introduces many spurious neighbors and makes edge reasoning harder.
Therefore, we set $K=8$ for Toulouse and $K=12$ for the other datasets, as the nodes in Toulouse road graphs contain fewer neighbors.
\textbf{4)} For the teacher-forcing decay length, a small $T_s$ substantially degrades performance because edge reasoning is exposed to predicted nodes before node detection becomes reliable, leading to corrupted candidate generation and edge supervision. Increasing $T_s$ improves training stability, but the gain from $T_s=30$ to $T_s=40$ is marginal. Therefore, we choose $T_s=30$, which provides stable early edge supervision while allowing edge reasoning to adapt to predicted nodes sufficiently early during training. 
Taken together, the selected settings provide a practical trade-off while achieving strong and stable performance across datasets.

\begin{table}[t]
\caption{Ablation study on the Synthetic dataset. We assess five variants: 1) w/o LineGT: replace line-graph transformer (Eq.~\ref{eq:line_GT}) with a standard graph transformer; 2) w/o ASNS: replace ASNS (Eq.~\ref{eq:asns}) with KNN; 3) w/o Visual: remove visual features (Eq.~\ref{eq:visual}); 4) w/o Spatial: remove spatial features (Eq.~\ref{eq:spatial}); 5) w/o Topology: remove global topology features (Eq.~\ref{eq:topology}).}
\label{tab:ablation}
\centering
\newcommand{\cmark}{\ding{51}}
\begin{tabular}{l@{\hskip 4pt}c@{\hskip 4pt}c@{\hskip 4pt}c@{\hskip 3.5pt}c}
\toprule
Variant &GIR$\uparrow$ & TOPO-F1$\uparrow$ & Edge-F1$\uparrow$  & Node-F1$\uparrow$ (\%)  \\
\midrule
Full &
\textbf{\meanstd{73.02}{0.46}} & \textbf{\meanstd{98.72}{0.43}} & \textbf{\meanstd{95.09}{1.25}} & \textbf{\meanstd{99.23}{0.37}}  \\
\midrule
w/o LineGT & \meanstd{54.43}{0.70}  & \meanstd{74.69}{0.38}  & \meanstd{72.63}{0.63}  & \meanstd{98.64}{0.25}    \\
w/o ASNS   & \meanstd{63.11}{0.68}  & \meanstd{86.47}{0.65}  & \meanstd{85.35}{0.94} & \meanstd{98.72}{0.41} \\
w/o Visual  & \meanstd{66.30}{0.57}   &  \meanstd{90.75}{0.35}  & \meanstd{89.36}{0.43} & \meanstd{98.82}{0.21} \\
w/o Spatial & \meanstd{70.26}{0.53} & \meanstd{95.68}{0.51} & \meanstd{93.81}{0.74} & \meanstd{99.07}{0.26}   \\
w/o Topology & \meanstd{68.59}{0.42}  & \meanstd{92.26}{0.81} & \meanstd{90.68}{0.86} & \meanstd{98.94}{0.19}  \\
\bottomrule
\end{tabular}
\end{table}

\subsection{Ablation Study}
We conduct ablation studies on the Synthetic dataset to assess the contribution of key components in the VisAdj framework.
Specifically, we evaluate five ablated variants, each modifying one component while keeping all other settings identical to the full model.

1) \textbf{w/o LineGT:} We replace the line-graph transformer (Eq.~\ref{eq:line_GT}) with a standard graph transformer \cite{ying2021transformers} that operates on node tokens. This variant removes the explicit modeling of edge-edge interactions among incident candidate edges and instead performs edge reasoning over node representations.

2) \textbf{w/o ASNS:} We replace the ASNS (Eq.~\ref{eq:asns}) with fixed KNN-based candidate selection, while keeping all other modules unchanged. This variant directly compares learned adaptive candidate selection with fixed KNN selection under identical edge reasoning conditions.
    
3) \textbf{w/o Visual:} We remove the visual features (Eq.~\ref{eq:visual}) sampled along candidate paths.
    
4) \textbf{w/o Spatial:} We remove the spatial features (Eq.~\ref{eq:spatial}) used for candidate edge representation.
    
5) \textbf{w/o Topology:} We remove the global topology-enhanced features (Eq.~\ref{eq:topology}) for candidate edge representation.

As shown in Table~\ref{tab:ablation}, each ablated variant results in a performance drop, confirming that the improvements of VisAdj rely on the coordinated design of all the proposed modules.

\textbf{1) Effect of edge-edge dependency modeling.}
Among all components, LineGT has the largest impact.
Replacing it with a standard graph transformer reduces GIR from $73.02\%$ to $54.43\%$ and Edge-F1 from $95.09\%$ to $72.63\%$.
This highlights the importance of explicitly modeling dependencies among incident candidate edges for globally consistent connectivity prediction.
It also shows that edge-level structured reasoning is more effective than relying only on node-level graph reasoning for adjacency reconstruction.

\textbf{2) Effect of adaptive candidate selection.}
ASNS also plays a critical role.
Replacing ASNS with fixed KNN candidate selection reduces GIR from $73.02\%$ to $63.11\%$ and TOPO-F1 from $98.72\%$ to $86.47\%$.
This demonstrates that learned candidate generation is important for capturing long-range or visually ambiguous connections and for overcoming the limitations of fixed KNN neighborhoods.

\textbf{3) Effect of edge representations.}
The remaining ablations validate the importance of visual, spatial, and topological features.
Removing visual features reduces GIR to $66.30\%$, showing that visual continuity cues along candidate paths help distinguish true edges from visually implausible connections.
Removing spatial features or global topology features also leads to performance degradation, lowering GIR to $70.26\%$ and $68.59\%$, respectively. This indicates that both geometric constraints and image-level structural context provide complementary signals for reliable edge reasoning.

\begin{figure}
    \centering
    \includegraphics[width=1.0\linewidth]{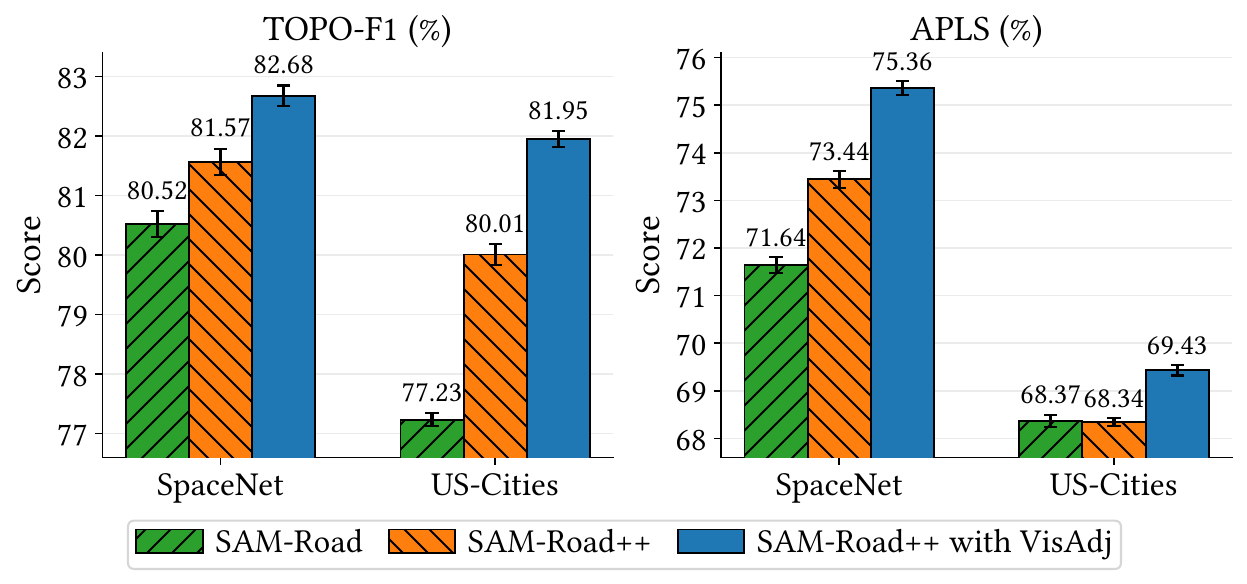}
    \caption{Road network extraction results on SpaceNet and US-Cities datasets. APLS is Average Path Length Similarity.}
    \Description{Comparison results for real-world road network extraction on SpaceNet and US-Cities datasets. }
    \label{fig:road_network_app}
\end{figure}

\subsection{Application in Road Network Extraction}
To evaluate practical applicability, we consider a real-world road network extraction scenario. 
This experiment follows the experimental protocol of SAM-Road++~\cite{yin2025towards} and assesses the effect of  VisAdj on an existing road extraction pipeline. 
Specifically, we integrate VisAdj into SAM-Road++ by replacing its graph reasoning module with VisAdj, while keeping all other components unchanged.  

As shown in Fig.~\ref{fig:road_network_app}, SAM-Road++ already improves over SAM-Road, forming a strong baseline. 
Integrating VisAdj into SAM-Road++ further improves both TOPO-F1 and APLS on SpaceNet~\cite{van2018spacenet} and US-Cities datasets. On US-Cities, TOPO-F1 increases from $80.01\%$ to $81.95\%$, and APLS improves from $68.34\%$ to $69.43\%$. 
On SpaceNet, VisAdj also improves TOPO-F1 from $81.57\%$ to $82.68\%$ and APLS from $73.44\%$ to $75.36\%$. 
These results show that VisAdj can be effectively integrated into real-world road extraction pipelines to improve both topological accuracy and path-level connectivity.

\section{Conclusion and Future Work}
In this work, we propose VisAdj, a unified framework for reconstructing adjacency matrices from node-link images. 
VisAdj combines adaptive candidate edge generation with explicit modeling of edge dependencies via a line-graph transformer, substantially improving adjacency reasoning beyond existing approaches that rely on fixed KNN candidate selection and predict each edge independently.
Future work will explore a permutation-invariant version \cite{xie2025advances} of VisAdj to decouple graph reasoning from node ordering and investigate broader application scenarios.
Another promising direction is to extend VisAdj to more general visual inputs where graph structures are implicitly embedded, such as scene graphs.


\begin{acks}
This project is supported in part by National Science Foundation under IIS-2144285 and IIS-2414308.
\end{acks}

\section*{GenAI Usage Disclosure}
During the preparation of this work, the authors used GenAI tools to help with code implementation and debugging.
No GenAI tools were used in the collection, processing, or annotation of the datasets.
GenAI tools were also used during the writing process to assist with grammar checking and sentence polishing.
All core concepts, methodologies, and experimental results were developed independently by the authors.
The authors have reviewed all GenAI-assisted content and take full responsibility for the accuracy and integrity of the work presented.

\bibliographystyle{ACM-Reference-Format}
\bibliography{sample-base}


\end{document}